\documentclass{article}

\usepackage[dblblindworkshop, final]{neurips_2026}
\workshoptitle{Foundation Models for Temporal Systems}

\usepackage[utf8]{inputenc}
\usepackage[T1]{fontenc}
\usepackage{times}
\usepackage{natbib}
\usepackage{amsmath,amssymb,amsfonts,amsthm,mathtools,bm}
\usepackage{booktabs}
\usepackage{graphicx}
\usepackage{microtype}
\usepackage{enumitem}
\usepackage{algorithm}
\usepackage{algpseudocode}
\usepackage[hidelinks]{hyperref}
\usepackage{url}
\usepackage{array}
\usepackage{tabularx}
\usepackage{multirow}
\usepackage{makecell}
\usepackage{float}

\title{\texttt{Sim+Real}: Joint Simulation--Experiment Training Improves Balanced Prediction in Physical Systems}

\author{%
  Mahindra Rautela, Alexander Scheinker, Ayan Biswas, Diane Oyen,\\ \textbf{Nathan Debardeleben}, \textbf{Earl Lawrence} \\
  Los Alamos National Laboratory, New Mexico, USA \\
  \texttt{mrautela@lanl.gov} \\
}

\newcommand{\Rsim}{R_{\mathrm{sim}}}
\newcommand{\Rexp}{R_{\mathrm{exp}}}

\newcommand{\Psim}{P_{\mathrm{sim}}}
\newcommand{\Pexp}{P_{\mathrm{exp}}}
\newcommand{\stageupdates}{10{,}000}

\begin{document}
\maketitle

\begin{abstract}
Simulation and experimental measurements provide complementary data for learning spatiotemporal physical systems, but standard simulation-to-experiment fine-tuning optimizes only the experimental objective after transfer and can degrade simulation performance. We formulate simulation--experiment prediction as a multi-objective learning problem with domain-specific simulation and experimental risks. On four fluid systems from RealPDEBench and two model capacities, we compare Simulation only, Experiment only, Sim$\rightarrow$Exp, and Joint training, evaluating every final model on both held-out domains. Sim$\rightarrow$Exp tends to specialize more strongly to experimental data at the cost of simulation-domain forgetting. Joint training consistently achieves the best balanced performance over a broad range of simulation--experiment evaluation weightings, while substantially improving simulation retention over Sim$\rightarrow$Exp. Joint also better preserves simulation-only fields absent from experimental measurements.
\noindent\textbf{Project page:} \url{https://mahindrautela.github.io/morph}
\end{abstract}

\section{Introduction}

Numerical simulations and real experiments provide complementary supervision for learning dynamics in physical systems. Simulations offer abundant, densely resolved trajectories and variables that may be inaccessible to measurement, but depend on modeling assumptions, constitutive laws, boundary conditions, discretization, and numerical solvers. Experiments observe the physical system directly, but are costly, noisy, and often only partially observed. The challenge is therefore to combine abundant but imperfect numerical supervision with limited experimental measurements.

Recently, RealPDEBench \citep{hu2026realpdebench} is released to handle this challenge by pairing numerical and experimental data across five physical systems and evaluating simulation-only training, experiment-only training, and simulation pretraining followed by experimental fine-tuning. The benchmark addresses the usual simulation-to-real objective: using simulation to improve experimental prediction. \emph{We ask a different question: if one model must remain useful on both domains, how should experimental adaptation be balanced against forgetting of simulation capability?}

We formulate joint simulation--experiment prediction as a \emph{multi-objective learning problem} with domain-specific simulation risk $\Rsim$ and experimental risk $\Rexp$ \citep{blitzer2007learning}. This separates two phenomena. The \emph{simulation-to-real distribution gap} is a discrepancy between $\Psim$ and $\Pexp$ that exists before adaptation. In contrast, we call an increase in held-out simulation error after experimental fine-tuning \emph{simulation-domain forgetting}. Sequential Sim$\rightarrow$Exp fine-tuning is appropriate when experimental risk $\Rexp$ is the only objective, but once fine-tuning begins, $\Rsim$ no longer constrains the parameter updates and simulation-domain forgetting
can occur. The resulting problem is a retention--adaptation trade-off rather than experimental adaptation alone.

We study balanced joint training (Joint), where one simulation mini-batch and one experimental mini-batch contribute to each optimizer update, corresponding to $\tfrac12\Rsim+\tfrac12\Rexp$. When the simulation and experimental gradients conflict, i.e., $g_{\mathrm{sim}}^\top g_{\mathrm{exp}}<0$, an experiment-only update can increase simulation error. Joint keeps a simulation-gradient contribution in every update, which helps preserve simulation performance while adapting to experimental data. Sequential and Joint training therefore represent different trade-offs between simulation retention and experimental adaptation.

Data heterogeneity between simulation and experiment introduces an additional architectural challenge for joint training. In the RealPDEBench fluid systems, simulations contain $(u,v,p)$, whereas particle-image-velocimetry measurements contain only $(u,v)$; the native spatial grids can also differ. This type of heterogeneity is increasingly addressed by PDE foundation models designed to operate across different physical fields and discretizations \citep{herde2024poseidon,mccabe2025walrus,holzschuh2025pdetransformer,rautela2025morph}. We therefore require a model that can process the fields available in each domain without discarding simulated variables or padding unobserved experimental fields. Field-wise cross-attention provides such an interface. Among the architectures we considered, MORPH \citep{rautela2025morph} provides this capability natively, so we use it at two model capacities and train both models from scratch on four RealPDEBench datasets. 

The architecture and datasets are not contributions of this work; they provide a controlled setting to study our central hypothesis: \emph{keeping simulation and experimental supervision active throughout training improves balanced performance across both domains and reduces simulation-domain forgetting compared with sequential Sim$\rightarrow$Exp fine-tuning.} To test this hypothesis, we compare Simulation only, Experiment only, Sim$\rightarrow$Exp, and Joint training using a shared architecture and protocol-matched optimization settings within each dataset and capacity. We evaluate two model capacities across four RealPDEBench fluid systems, testing every trained model on both simulation and experimental data. To summarize performance across the two domains, we define equal-domain balanced RMSE, relative $L_2$, and Fourier RMSE. Our claim is therefore specifically about \emph{balanced dual-domain performance}, rather than universal superiority on either domain individually.

Our contributions are:
\begin{enumerate}
\item We formulate simulation--experiment PDE prediction as a multi-objective learning problem with domain-specific simulation and experimental risks.

\item We compare Simulation only, Experiment only, Sim$\rightarrow$Exp, and Joint training with matched initialization and optimization settings within each dataset and capacity.

\item We evaluate both domains across four RealPDEBench fluid systems and two model capacities using three balanced dual-domain metrics.

\end{enumerate}

\section{Related Work}
\label{app:related_works}

\paragraph{Learning PDE dynamics.}
Data-driven models for physical dynamics span continuous-time models, neural operators, and reduced-order latent representations. Neural ODEs learn continuous-time dynamics directly from data \citep{chen2018neural}, while DeepONet \citep{goswami2022deep} and FNO \citep{li2020fourier} learn mappings between function spaces for PDE solution operators. Latent evolution models instead compress high-dimensional physical states and learn their dynamics in a lower-dimensional representation \citep{maulik2021reduced,rautela2025time,solera2024beta}. More recently, PDE foundation models such as DPOT \citep{hao2024dpot}, Poseidon \citep{herde2024poseidon}, Walrus \citep{mccabe2025walrus}, PDE-Transformer \citep{holzschuh2025pdetransformer}, and MORPH \citep{rautela2025morph} target transfer across PDE systems, physical variables, and discretizations. Our work does not introduce a new architecture; we study how such models should be trained when numerical and experimental data are both available.

\paragraph{Simulation-to-real transfer in physics.}
A common strategy for data-scarce physical systems is to learn from abundant simulations and subsequently adapt the model using limited observations from the physical system. Transfer learning has been used to calibrate simulation-trained surrogate models to experimental data in inertial-confinement fusion \citep{humbird2020transfer}. Related work also studies efficient adaptation of learned dynamics across changing physical systems \citep{kirchmeyer2022generalizing}. RealPDEBench directly formalizes simulation-to-real PDE prediction using paired numerical and experimental datasets and evaluates simulation pretraining followed by experimental fine-tuning \citep{hu2026realpdebench}. Recent work further shows that physics foundation models trained on simulations can transfer to laboratory turbulence \citep{mukhopadhyay2026emergent}. In contrast, we ask what happens to simulation performance after experimental adaptation when the final model is expected to remain useful on both domains.

\paragraph{Domain forgetting and joint optimization.}
Sequential adaptation can degrade performance on previously learned data, a phenomenon broadly studied as catastrophic forgetting in continual learning \citep{lopezpaz2017gradient,bai2024continual}. Gradient Episodic Memory constrains updates to avoid increasing loss on previous tasks \citep{lopezpaz2017gradient}, while domain-adaptation theory studies the trade-off between source and target empirical risks \citep{blitzer2007learning}. Closely related optimization work formulates learning with competing objectives as multi-objective optimization \citep{sener2018multi} and shows that conflicting gradients can cause detrimental interference between objectives \citep{yu2020gradient}. Our setting differs from conventional continual learning because simulation data remain available during adaptation. We use this availability to compare sequential Sim$\rightarrow$Exp fine-tuning with Joint training that keeps both domain objectives active.

\section{Proposed Method}
\label{app:methods}

\subsection{Problem Setup}

Let $d\in\{\mathrm{sim},\mathrm{exp}\}$ denote the data domain, with distribution $\mathcal{P}_d$ and training dataset
\begin{equation}
    \mathcal{D}_d
    =
    \{(\bm{x}_i^d,\bm{y}_i^d)\}_{i=1}^{N_d},
    \qquad
    (\bm{x}^d,\bm{y}^d)\sim\mathcal{P}_d,
\end{equation}
where $N_d$ is the number of training examples, $\bm{x}_i^d$ is the
input time window, and $\bm{y}_i^d$ is the supervised prediction target.

Each domain is characterized by an input field set
$\mathcal{F}_{\mathrm{in}}^d$, a supervised output field set
$\mathcal{F}_{\mathrm{out}}^d\subseteq\mathcal{F}_{\mathrm{in}}^d$,
and a spatial grid $\mathcal{G}^d$. The available fields differ between
simulation and experiment,
\begin{equation}
    \mathcal{F}_{\mathrm{in}}^{\mathrm{sim}}
    \neq
    \mathcal{F}_{\mathrm{in}}^{\mathrm{exp}},
\end{equation}
and the spatial grids $\mathcal{G}^{\mathrm{sim}}$ and
$\mathcal{G}^{\mathrm{exp}}$ may also differ.

For the four fluid systems considered in the main study, the supervised
dynamic fields are
\begin{equation}
    \mathcal{F}_{\mathrm{out}}^{\mathrm{sim}}=\{u,v,p\},
    \qquad
    \mathcal{F}_{\mathrm{out}}^{\mathrm{exp}}=\{u,v\}.
\end{equation}
Controlled Cylinder additionally includes static conditioning fields in
$\mathcal{F}_{\mathrm{in}}^d$ that are not prediction targets.

For a shared model $f_\theta$, define the per-example prediction
loss in domain $d$ as
\begin{equation}
    \ell_d(\theta;\bm{x}^d,\bm{y}^d)
    =
    \frac{1}
    {|\mathcal{F}_{\mathrm{out}}^d|\,|\mathcal{G}^d|}
    \sum_{f\in\mathcal{F}_{\mathrm{out}}^d}
    \sum_{j\in\mathcal{G}^d}
    \left(
        f_\theta(\bm{x}^d)_{f,j}
        -
        y_{f,j}^d
    \right)^2,
\end{equation}
where $j$ indexes spatial locations. The corresponding domain-specific
population risk is
\begin{equation}
    R_d(\theta)
    =
    \mathbb{E}_{(\bm{x}^d,\bm{y}^d)\sim\mathcal{P}_d}
    \left[
        \ell_d(\theta;\bm{x}^d,\bm{y}^d)
    \right].
\end{equation}
Its finite-data counterpart is
\begin{equation}
    \widehat{R}_d(\theta)
    =
    \frac{1}{N_d}
    \sum_{i=1}^{N_d}
    \ell_d(\theta;\bm{x}_i^d,\bm{y}_i^d),
\end{equation}
which is estimated by mini-batches during training. Averaging over both supervised fields and spatial locations prevents a domain from receiving a larger loss solely because it contains more fields or a higher-resolution grid.

The simulation and experimental risks need not have the same minimizer. The learning problem is therefore naturally multi-objective:
\begin{equation}
    \min_\theta
    \left(
        R_{\mathrm{sim}}(\theta),
        R_{\mathrm{exp}}(\theta)
    \right).
    \label{eq:multiobjective}
\end{equation}

Training uses all supervised dynamic fields available in each domain,
whereas the primary dual-domain evaluation for the four fluid systems is
restricted to the common measured fields $(u,v)$.

\subsection{Proposed Modeling Requirement for Heterogeneous Data}

The optimization study requires one model to process samples whose field sets differ across domains. We therefore represent each available physical field independently. For field $f\in\mathcal{F}^d$, let
\begin{equation}
    \bm{Z}_f^d
    =
    P(\bm{x}_f^d;\mathcal{G}^d)+\bm{e}_f,
    \label{eq:field_tokens}
\end{equation}
where $P$ patchifies the field on its native grid and $\bm{e}_f$ identifies the physical variable. The available field tokens are concatenated,
\begin{equation}
    \bm{Z}^d
    =
    \operatorname{concat}_{f\in\mathcal{F}^d}\bm{Z}_f^d,
\end{equation}
and a fixed latent query set $\bm{Q}$ fuses this variable-size collection through cross-attention:
\begin{equation}
    \bm{H}^d
    =
    \operatorname{softmax}\!\left(
    \frac{(\bm{Q}W_Q)(\bm{Z}^dW_K)^\top}{\sqrt{d_m}}
    \right)\bm{Z}^dW_V.
    \label{eq:field_cross_attention}
\end{equation}
The prediction backbone therefore receives a fixed latent representation even when the number and identity of observed physical fields change. Unavailable experimental fields are omitted rather than zero-padded, while simulation samples retain their additional fields. No explicit simulation/experiment embedding is added in our implementation.

\subsection{Sequential and Joint Objectives}

The two domain risks define a bi-objective optimization problem. Joint
training uses the linear scalarization
\begin{equation}
    R_{\lambda}(\theta)
    =
    \lambda R_{\mathrm{sim}}(\theta)
    +
    (1-\lambda)R_{\mathrm{exp}}(\theta),
    \qquad
    \lambda=\tfrac12 .
    \label{eq:joint}
\end{equation}

For a mini-batch $B_d$ from domain $d$, define
\begin{equation}
    \widehat{R}_d(\theta;B_d)
    =
    \frac{1}{|B_d|}
    \sum_{(\bm{x},\bm{y})\in B_d}
    \ell_d(\theta;\bm{x},\bm{y}).
\end{equation}
A Joint update uses the stochastic gradient
\begin{equation}
    \widehat g_{\mathrm{joint}}
    =
    \lambda\nabla_\theta\widehat R_{\mathrm{sim}}(\theta;B_s)
    +
    (1-\lambda)\nabla_\theta\widehat R_{\mathrm{exp}}(\theta;B_e),
\end{equation}
which is an unbiased estimator of $\nabla_\theta R_\lambda(\theta)$
under independent mini-batch sampling.

\subsection{Why sequential fine-tuning can forget}
\label{sec:gradient_analysis}

To expose the local mechanism behind simulation-domain forgetting, consider a single gradient-descent step before optimizer preconditioning and define
\begin{equation}
    g_s = \nabla_\theta R_{\mathrm{sim}}(\theta),
    \qquad
    g_e = \nabla_\theta R_{\mathrm{exp}}(\theta).
    \label{eq:domain_gradients}
\end{equation}
For a small parameter displacement $\Delta\theta$, a first-order Taylor expansion of either domain risk gives
\begin{equation}
    R_d(\theta+\Delta\theta)
    =
    R_d(\theta)
    +
    \nabla_\theta R_d(\theta)^\top \Delta\theta
    +
    \mathcal{O}(\|\Delta\theta\|_2^2),
    \qquad d\in\{\mathrm{sim},\mathrm{exp}\}.
    \label{eq:taylor_risk}
\end{equation}

During the experimental stage of Sim$\rightarrow$Exp, the update is
\begin{equation}
    \theta^+ = \theta - \eta g_e,
    \label{eq:seq_update}
\end{equation}
so that $\Delta\theta=-\eta g_e$. Applying Eq.~\eqref{eq:taylor_risk} to the simulation risk therefore gives
\begin{align}
    \Delta R_s^{\mathrm{seq}}
    &\equiv
    R_{\mathrm{sim}}(\theta^+) - R_{\mathrm{sim}}(\theta)
    \nonumber\\
    &=
    g_s^\top(-\eta g_e)
    +
    \mathcal{O}(\eta^2\|g_e\|_2^2)
    \nonumber\\
    &=
    -\eta g_s^\top g_e
    +
    \mathcal{O}(\eta^2\|g_e\|_2^2).
    \label{eq:seq_sim_change}
\end{align}
The inner product $g_s^\top g_e$ measures local gradient alignment. If
$g_s^\top g_e<0$, the two gradients point in conflicting directions and the
leading-order term in Eq.~\eqref{eq:seq_sim_change} is positive. Hence, for a
sufficiently small step, an update that decreases experimental risk can
increase simulation risk, providing a direct local mechanism for
simulation-domain forgetting.

For Joint training, the scalarized objective gives the gradient
\begin{equation}
    g_{\mathrm{joint}}
    =
    \lambda g_s + (1-\lambda)g_e,
    \label{eq:joint_gradient}
\end{equation}
and the corresponding update is
$\theta^+=\theta-\eta g_{\mathrm{joint}}$. Expanding the simulation risk in
this direction yields
\begin{align}
    \Delta R_s^{\mathrm{joint}}
    &=
    -\eta g_s^\top g_{\mathrm{joint}}
    +
    \mathcal{O}(\eta^2\|g_{\mathrm{joint}}\|_2^2)
    \nonumber\\
    &=
    -\eta
    \left[
        \lambda\|g_s\|_2^2
        +
        (1-\lambda)g_s^\top g_e
    \right]
    +
    \mathcal{O}(\eta^2\|g_{\mathrm{joint}}\|_2^2).
    \label{eq:joint_sim_change}
\end{align}
Similarly, the change in experimental risk is
\begin{align}
    \Delta R_e^{\mathrm{joint}}
    &=
    -\eta g_e^\top g_{\mathrm{joint}}
    +
    \mathcal{O}(\eta^2\|g_{\mathrm{joint}}\|_2^2)
    \nonumber\\
    &=
    -\eta
    \left[
        (1-\lambda)\|g_e\|_2^2
        +
        \lambda g_s^\top g_e
    \right]
    +
    \mathcal{O}(\eta^2\|g_{\mathrm{joint}}\|_2^2).
    \label{eq:joint_exp_change}
\end{align}

Equations~\eqref{eq:joint_sim_change}--\eqref{eq:joint_exp_change} separate
each update into a domain-specific descent term and a cross-domain
interaction term. In Eq.~\eqref{eq:joint_sim_change},
$-\eta\lambda\|g_s\|_2^2$ always favors decreasing simulation risk and
therefore counteracts a conflicting experimental gradient. Joint does not
guarantee that simulation risk decreases under arbitrary conflict: to first
order, this requires
\begin{equation}
    \lambda\|g_s\|_2^2
    +
    (1-\lambda)g_s^\top g_e > 0.
    \label{eq:joint_sim_condition}
\end{equation}
The symmetric expression for experimental risk shows the corresponding retention--adaptation trade-off. When $g_s^\top g_e<0$, preserving simulation capability can reduce the rate of experimental specialization,
whereas sequential fine-tuning contains no simulation-descent term once the experimental stage begins. This local analysis therefore motivates the empirical behavior tested below: Sim$\rightarrow$Exp may specialize more strongly to experimental data, while Joint can retain substantially more simulation capability by keeping both objectives active.

\section{Experimental Setup}

\textbf{Datasets.}
We use the four RealPDEBench fluid systems \citep{hu2026realpdebench}. Table~\ref{tab:datasets} reports the native benchmark fields and grids. All four systems share measured velocity fields and form the dual-domain study. For Cylinder, FSI, and Foil, the input and target blocks contain 20 frames; Controlled Cylinder uses 10-frame input and target blocks. We use the public benchmark splits and the Hugging Face dataset representation. Numerical-training augmentation follows the benchmark configuration; validation, test, and normalization datasets are clean. All protocols use Gaussian normalization statistics computed from the numerical training split, matching the implemented RealPDEBench pipeline. 

\begin{table}[t]
\centering
\caption{RealPDEBench systems. The four fluid datasets share velocity observables $(u,v)$; simulation additionally contains pressure. The grids shown are the native benchmark grids.}
\label{tab:datasets}
\small
\setlength{\tabcolsep}{4pt}
\begin{tabular}{lcccc}
\toprule
Dataset & Sim. fields & Exp. fields & Sim. grid & Exp. grid \\
\midrule
Cylinder & $u,v,p$ & $u,v$ & $64\!\times\!128$ & $128\!\times\!256$ \\
Controlled Cylinder & $u,v,p$ & $u,v$ & $64\!\times\!128$ & $128\!\times\!256$ \\
FSI & $u,v,p$ & $u,v$ & $128\!\times\!128$ & $128\!\times\!128$ \\
Foil & $u,v,p$ & $u,v$ & $128\!\times\!256$ & $128\!\times\!256$ \\
\bottomrule
\end{tabular}
\end{table}

\paragraph{Architecture.}
Joint training requires one model to process simulation and experimental
samples with different field sets and, in some systems, different spatial
grids. We use MORPH \citep{rautela2025morph}, whose field-wise representation
supports these heterogeneous inputs without padding missing experimental fields
or discarding additional simulation fields. Architecture comparisons and the
full modeling interface are provided in Table~\ref{tab:architecture_selection} in Appendix~\ref{app:reproducibility}.

\textbf{Prediction protocol.}
We retain the RealPDEBench temporal windows: 20 input and 20 target frames for Cylinder, FSI, and Foil, and 10 input and 10 target frames for Controlled Cylinder. MORPH is trained as a next-state predictor, where a single future frame is predicted from a fixed-length teacher-forced context. At test time, predictions are generated autoregressively by feeding each predicted state back into the context until the full target horizon is reached. \emph{We adopt this frame-wise autoregressive setting because practical forecasting requires evolving the physical system without access to future ground-truth states, directly exposing error accumulation and long-horizon stability.} This differs from the standard RealPDEBench baselines, which predict the complete target block in one round and autoregress only across predicted blocks. Therefore, MORPH results should be interpreted under a frame-wise autoregressive protocol rather than as an architecture-matched reproduction of the RealPDEBench baselines.

\textbf{Models and optimization.}
After the capability selection above, we instantiate MORPH from scratch (no pretrained weights) at two capacities: 17.5M and 48.2M trainable parameters. These correspond to the Ti and S configurations with patch size 4 and a maximum of 2,048 patches. For each dataset and capacity, the random seed is reset before model construction so all four training modes start from the same initialization. The architecture is held fixed across training protocols because the experimental question concerns the training protocol. All experiments were conducted using four NVIDIA A100 GPUs. All final runs use Adam with learning rate $10^{-4}$, cosine annealing over each $\stageupdates$-update stage, no gradient accumulation, no gradient clipping, and automatic mixed precision. The physical training batch size is 6 for Cylinder, Controlled Cylinder, and FSI and 3 for Foil; within each dataset, the same batch size is used for all four training protocols. Evaluation uses batch size 1 throughout. Joint uses weights $\lambda=1/2$ for both sources. Full data-pipeline, optimization, checkpointing, and evaluation details are given in Appendix~\ref{app:reproducibility}. 

\textbf{Evaluation and balanced dual-domain metrics.}
Every fluid checkpoint is evaluated on both held-out simulation and experimental test sets. To make the two domains directly comparable, metrics are computed only on the shared velocity fields $(u,v)$. We report the RealPDEBench RMSE, relative $L_2$, and fRMSE metrics \citep{hu2026realpdebench,takamoto2022pdebench}; full domain-wise results and evaluation implementation details are provided in Appendices~\ref{app:reproducibility} and \ref{app:detailed_results}. Because Joint optimizes an equal-weight dual-domain objective, we summarize performance across simulation and experiment with
\begin{equation}
\mathrm{bRMSE}=\sqrt{\frac{E_s^2+E_e^2}{2}}, \qquad
\mathrm{bRelL2}=\frac{\mathrm{RelL2}_{\mathrm{sim}}+\mathrm{RelL2}_{\mathrm{exp}}}{2}, \qquad
\mathrm{bfRMSE}=\sqrt{\frac{F_s^2+F_e^2}{2}},
\label{eq:balanced_metrics}
\end{equation}
where $E_s,E_e$ and $F_s,F_e$ denote simulation and experimental RMSE and fRMSE, respectively. All three summaries assign equal weight to the two domains ($\lambda=1/2$), independently of test-set size. They measure balanced physical-space, relative, and Fourier-space error, respectively, while the raw domain-wise metrics are reported in Appendix~\ref{app:detailed_results}. 

\section{Results}

\subsection{Joint training gives the best balanced dual-domain performance.}
Table~\ref{tab:balanced_dual_domain} is the main result. Joint achieves the lowest bRMSE, bRelL2, and bfRMSE in \emph{all eight} fluid dataset--model settings, giving 24/24 wins across the three balanced metrics. Relative to the best non-joint protocol in each row, Joint reduces bRMSE by 39.0--58.4\%, with a mean improvement of 49.1\%. For bRelL2, the improvement is 35.2--68.5\%, with a mean of 52.4\%. For bfRMSE, the improvement is 38.1--66.4\%, with a mean of 53.0\%. The agreement of physical-space RMSE, dimensionless relative $L_2$, and Fourier-space RMSE is important: the balanced advantage is not tied to one error representation or to the scale of a single metric.

Appendix~\ref{app:lambda} further shows that this advantage is not specific to equal-domain evaluation: with the trained checkpoints fixed, Joint is best in all 24 dataset--capacity--metric comparisons at every sampled evaluation weight $\lambda\in\{0.2,0.3,\ldots,0.8\}$, and remains best in 22/24 and 21/24 comparisons at $\lambda=0.1$ and $\lambda=0.9$, respectively.

\begin{table}[t]
\centering
\caption{\textbf{Main result: balanced dual-domain performance} on the four fluid datasets. bRMSE, bRelL2, and bfRMSE give simulation and experiment equal weight according to Eq.~\eqref{eq:balanced_metrics}. Lower is better. Bold marks the best protocol for each dataset and model capacity. Joint is best in all 24 comparisons (8 settings $\times$ 3 metrics).}
\label{tab:balanced_dual_domain}
\scriptsize
\setlength{\tabcolsep}{2.6pt}
\resizebox{\textwidth}{!}{%
\begin{tabular}{ll ccc ccc ccc ccc}
\toprule
& &
\multicolumn{3}{c}{Simulation only} &
\multicolumn{3}{c}{Experiment only} &
\multicolumn{3}{c}{Sim$\rightarrow$Exp} &
\multicolumn{3}{c}{Joint} \\
\cmidrule(lr){3-5}\cmidrule(lr){6-8}\cmidrule(lr){9-11}\cmidrule(lr){12-14}
Dataset & Capacity
& bRMSE & bRelL2 & bfRMSE
& bRMSE & bRelL2 & bfRMSE
& bRMSE & bRelL2 & bfRMSE
& bRMSE & bRelL2 & bfRMSE \\
\midrule
Cylinder & 17.5M
& 0.118098 & 0.539958 & 0.019925
& 0.101539 & 0.332502 & 0.015969
& 0.100507 & 0.419261 & 0.015714
& \textbf{0.058412} & \textbf{0.126032} & \textbf{0.009489} \\
Cylinder & 48.2M
& 0.096126 & 0.427966 & 0.015923
& 0.094315 & 0.311948 & 0.014704
& 0.094171 & 0.354185 & 0.014653
& \textbf{0.056230} & \textbf{0.098416} & \textbf{0.009077} \\
\midrule
Ctrl Cyl & 17.5M
& 0.049899 & 0.333784 & 0.012611
& 0.061036 & 0.405474 & 0.014763
& 0.071465 & 0.490167 & 0.017985
& \textbf{0.030442} & \textbf{0.216437} & \textbf{0.006148} \\
Ctrl Cyl & 48.2M
& 0.048514 & 0.298634 & 0.012441
& 0.058852 & 0.393601 & 0.014363
& 0.077844 & 0.481965 & 0.019656
& \textbf{0.020455} & \textbf{0.149020} & \textbf{0.004180} \\
\midrule
FSI & 17.5M
& 0.060337 & 0.405078 & 0.009692
& 0.082525 & 0.463016 & 0.013427
& 0.069920 & 0.392799 & 0.010924
& \textbf{0.033517} & \textbf{0.229200} & \textbf{0.004630} \\
FSI & 48.2M
& 0.070413 & 0.456217 & 0.011576
& 0.073238 & 0.402001 & 0.011503
& 0.063963 & 0.345480 & 0.009488
& \textbf{0.028268} & \textbf{0.198546} & \textbf{0.003761} \\
\midrule
Foil & 17.5M
& 0.094142 & 0.341876 & 0.014331
& 0.077088 & 0.280055 & 0.011590
& 0.063129 & 0.246719 & 0.010051
& \textbf{0.026247} & \textbf{0.094245} & \textbf{0.003950} \\
Foil & 48.2M
& 0.077296 & 0.276485 & 0.011891
& 0.086083 & 0.318176 & 0.012722
& 0.078401 & 0.304760 & 0.011831
& \textbf{0.034927} & \textbf{0.117081} & \textbf{0.005293} \\
\bottomrule
\end{tabular}}
\end{table}

\subsection{Sequential fine-tuning specializes to experiment and forgets simulation.}

The detailed results in Appendix~\ref{app:detailed_results} expose the mechanism behind Table~\ref{tab:balanced_dual_domain}. Sim$\rightarrow$Exp has higher simulation RMSE than Simulation only in all eight fluid settings. The relative increase ranges from 150.9\% to 608.6\%, with a median of 335.5\%. Joint, in contrast, has lower simulation RMSE than Sim$\rightarrow$Exp in all eight settings, reducing the sequential model's simulation RMSE by 60.0--87.2\% (median 66.9\%). This is direct test-set evidence of simulation-domain forgetting under sequential adaptation and of substantially better retention under Joint.

The price of retention is also visible. Sim$\rightarrow$Exp gives the lowest experimental RMSE in six of the eight fluid settings, while Joint is best in the two Cylinder settings. Thus, when the two objectives compete, sequential fine-tuning can specialize more aggressively to experiment, consistent with Eqs.~\eqref{eq:taylor_risk}, \eqref{eq:joint_sim_change} \& \eqref{eq:joint_exp_change}. The central empirical claim is therefore about \emph{balanced dual-domain performance}, not universal dominance on the experimental domain.

The raw-domain results also show that Joint is not merely interpolating between two domain-specific optima. Simulation only gives the lowest simulation RMSE in six of eight settings, while Joint is best for Cylinder-S and FSI-S. Conversely, Joint gives the lowest experimental RMSE for both Cylinder capacities, whereas Sim$\rightarrow$Exp is best for both capacities of Controlled Cylinder, FSI, and Foil. The balanced objective therefore selects a model that remains useful on both domains, while the single-domain winners depend on the dataset and capacity.

Importantly, this retention gap also appears for pressure, which is supervised in simulation but absent from the experimental measurements. Appendix~\ref{app:pressure_retention} shows that Joint reduces pressure RMSE by 52.4--86.6\% relative to Sim$\rightarrow$Exp, indicating substantially better preservation of this simulation-only field during experimental adaptation.

\section{Conclusion}

Simulation-to-experiment fine-tuning and joint simulation--experiment learning optimize different goals. Sequential fine-tuning produces the lowest experimental RMSE in six of eight settings, but increases simulation RMSE relative to simulation-only training in every dataset--model setting. Joint training keeps both objectives active, substantially reduces this forgetting, and achieves the best equal-domain balanced RMSE, relative $L_2$, and Fourier RMSE in all eight settings. The gradient analysis provides a simple explanation: experimental updates can increase simulation risk when domain gradients conflict, whereas Joint retains a simulation-gradient term and therefore continues to optimize both risks. 

\section*{Acknowledgements}
Research presented in this article was supported by the Laboratory Directed Research and Development program of Los Alamos National Laboratory under project number 20250637DI. This research used resources provided by the Los Alamos National Laboratory Institutional Computing Program, which is supported by the U.S. Department of Energy National Nuclear Security Administration under Contract No. 89233218CNA000001

\bibliographystyle{plainnat}
\bibliography{references}

\newpage
\appendix

\section{Experimental details}
\label{app:reproducibility}

This section records the settings used by the final experiment code and configurations: \texttt{p4\_b6\_10k} for Cylinder, Controlled Cylinder, and FSI, and \texttt{p4\_b3\_10k} for Foil. The intent is to make the comparison reproducible and to make explicit which quantities are held fixed across training protocols.

\subsection{Dataset construction and preprocessing}

We instantiate the public RealPDEBench Hugging Face dataset classes for Cylinder, Controlled Cylinder, FSI, and Foil, using the benchmark train/validation/test index files with \texttt{test\_mode=all}. Training loaders are shuffled; validation and test loaders are deterministic. The physical training batch size is 6 for Cylinder, Controlled Cylinder, and FSI and 3 for Foil; the validation and test batch size is 1 for every dataset. Within a dataset, the training batch size is identical across Simulation only, Experiment only, Sim$\rightarrow$Exp, and Joint. Each loader uses four worker processes. No test sample is used during optimization or checkpoint selection.

Table~\ref{tab:dataset_training_config} gives the temporal windows and numerical-training augmentation used in the final configurations. Numerical augmentation is enabled only for the numerical \emph{training} loader. The experimental training loader, both validation/test loaders, and the numerical dataset used to construct normalization statistics are unaugmented. The noise type is Gaussian whenever the noise scale is nonzero.

\begin{table}[H]
\centering
\caption{Dataset-specific temporal and numerical-training settings used by the final runs. Masking and noise are applied only by the RealPDEBench numerical training dataset; validation, test, experimental training, and normalization data are clean.}
\label{tab:dataset_training_config}
\small
\setlength{\tabcolsep}{5pt}
\begin{tabular}{lcccc}
\toprule
Dataset & Input frames & Target frames & Mask probability & Noise scale \\
\midrule
Cylinder & 20 & 20 & 0.1 & 0.1 \\
Controlled Cylinder & 10 & 10 & 0.1 & 0.1 \\
FSI & 20 & 20 & 0.5 & 0.1 \\
Foil & 20 & 20 & 0.1 & 0.1 \\
\bottomrule
\end{tabular}
\end{table}

All protocols use the same Gaussian normalizer. Its statistics are computed from a clean copy of the \emph{numerical training split} with normalizer batch size 512 and are then reused for simulation and experimental samples in every training mode. Consequently, ``Experiment only'' refers to the source of supervised updates; it does not use separately fitted experimental normalization statistics.

The repository does not hard-code the number of temporal windows in each split: those counts are determined by the public RealPDEBench index files and are written at runtime to \texttt{data\_summary.json}. We therefore avoid duplicating potentially artifact-dependent window counts in the manuscript and specify the split construction instead.

\subsection{Field mapping and heterogeneous inputs}

RealPDEBench returns tensors in $[B,T,H,W,C]$ format. Each available channel is treated as one scalar physical field and is reshaped to the field-wise representation $[B,T,F,1,1,H,W]$ expected by the selected cross-attention interface. No placeholder field is passed through the model for an unavailable experimental observable.

For Cylinder, Controlled Cylinder, FSI, and Foil, numerical samples use dynamic channels $(u,v,p)$, whereas experimental samples use only $(u,v)$. Controlled Cylinder additionally appends two static condition maps after the dynamic channels; these maps are repeated across the input context and held fixed during autoregressive rollout. The model is instantiated with a maximum of 20 fields for every dataset, so model capacity does not change with the observed field count.

\subsection{Forecast target construction and autoregressive evaluation}
The selected decoder predicts one future state. During training, let $T_{\mathrm{out}}$ be the number of target frames in Table~\ref{tab:dataset_training_config}. For each training sample, the implementation draws
\begin{equation}
    h \sim \mathrm{Unif}\{0,\ldots,T_{\mathrm{out}}-1\}
\end{equation}
and forms a one-step teacher-forced pair. The context consists of the preceding \texttt{context\_steps} states from the concatenation of the input block and the ground-truth target block, and the target is the single state at horizon $h$. The training loss is mean-squared error over the dynamic fields available in that domain. Thus, simulation training uses $(u,v,p)$ for the four fluid systems, while experimental training uses $(u,v)$.

Validation and test use a different rollout protocol: the model predicts one state at a time, appends that prediction to the history, and repeats until the complete benchmark target block has been generated. Controlled Cylinder uses 10 context steps; all other datasets use 20. The maximum temporal embedding length remains 20 for every dataset, so changing the Controlled Cylinder context length does not change the parameter count.

\subsection{Architecture requirements and selection.}
The architecture is selected only to enable the controlled simulation--experiment training study. A single model must process samples from both domains without changing its backbone, even though the two sources can differ in spatial resolution and in the physical fields that are available. For our study, we specifically use a field-wise cross-attention interface. The prediction backbone therefore receives a common latent representation without zero-padding missing experimental fields or discarding additional simulation fields. Table~\ref{tab:architecture_selection} summarizes the interface capabilities of representative architectures. Among the models considered, MORPH \citep{rautela2025morph} natively provides the combination of variable-grid processing, variable field sets, and field-wise cross-attention used in our implementation. We therefore select it as the fixed enabling architecture for all training protocols. Architecture design itself is not the contribution of this work.

\begin{table}[t]
\centering
\caption{
Capability-based architecture selection for heterogeneous
simulation--experiment training.
``Partial'' denotes that padding, interpolation, resizing, or
task-specific interfaces are required in the standard formulation.
The table compares interface capabilities rather than predictive accuracy.
}
\label{tab:architecture_selection}
\small
\setlength{\tabcolsep}{4pt}
\resizebox{\textwidth}{!}{%
\begin{tabular}{lcccc}
\toprule
Architecture
& Variable grid
& Variable field set
& Field-wise cross-attention
& Selected \\
\midrule

Vanilla ViT \citep{dosovitskiy2021image}
& Partial
& No
& No
& No \\

U-Net \citep{ronneberger2015unet}
& Partial
& No
& No
& No \\

FNO \citep{li2021fno}
& Yes
& No
& No
& No \\

DPOT \citep{hao2024dpot}
& Partial
& Partial
& No
& No \\

Poseidon \citep{herde2024poseidon}
& Yes
& Partial
& No
& No \\

Walrus \citep{mccabe2025walrus}
& Yes
& Partial
& No
& No \\

PDE-Transformer \citep{holzschuh2025pdetransformer}
& Yes
& Yes
& No
& No \\

MORPH \citep{rautela2025morph}
& Yes
& Yes
& Yes
& Yes \\

\bottomrule
\end{tabular}}
\end{table}

\subsection{MORPH Architecture and Prior Applications}

MORPH is a modality-agnostic, autoregressive PDE foundation model designed to handle heterogeneous scientific data across 1D--3D domains, varying resolutions, and mixed scalar and vector fields \citep{rautela2025morph}.
Its architecture combines component-wise convolutions for local field interactions, inter-field cross-attention for information exchange across physical variables, and factorized axial attention over spatial and temporal
dimensions. A Unified Physics Tensor Format (UPTF-7) provides a common representation across heterogeneous datasets. Beyond forward prediction, MORPH has been adapted to inverse parameter estimation in inertial confinement
fusion using multimodal hyperspectral and scalar observations \citep{rautela2026pde}, and evaluated for out-of-distribution transfer to shock-driven perturbed layered interfaces (PLI) and dynamic fracture/failure
(FRAC) \citep{rautela2026out}. These capabilities make MORPH a suitable backbone for the heterogeneous simulation--experiment setting considered here.

\subsection{Architecture instantiation}

The same field-wise cross-attention implementation is used for every training protocol. Table~\ref{tab:architecture_hparams} records the two capacities. In addition to the size-specific parameters, both use patch size 4, 32 cross-attention heads at the field-fusion interface, maximum 20 fields, maximum 2,048 patches, maximum 20 context steps, one component per field, dropout 0.1, embedding dropout 0.1, and no LoRA adapters. No released pretrained weights are loaded.

\begin{table}[H]
\centering
\caption{Architecture configurations used in the controlled study. Parameter counts are the trainable counts of the instantiated models.}
\label{tab:architecture_hparams}
\small
\setlength{\tabcolsep}{5pt}
\begin{tabular}{lrrrrrr}
\toprule
Configuration & Parameters & Conv. filters & Latent dim. & Heads & Depth & MLP dim. \\
\midrule
Ti & 17,528,776 & 8 & 256 & 4 & 4 & 1024 \\
S  & 48,161,736 & 8 & 512 & 8 & 4 & 2048 \\
\bottomrule
\end{tabular}
\end{table}

\subsection{Source-exposure-matched comparison}
The final experiments use a stage budget $U=\stageupdates$. Let $B_d$ denote the physical training batch size for dataset $d$: $B_d=6$ for Cylinder, Controlled Cylinder, and FSI, and $B_d=3$ for Foil. Simulation only and Experiment only use $U$ optimizer updates from one source. Sim$\rightarrow$Exp uses $U$ simulation updates followed by $U$ experimental updates; the optimizer and cosine scheduler are reset before the experimental stage. Joint uses $U$ optimizer updates, but each update processes one simulation and one experimental mini-batch. Therefore, within each dataset, Joint and Sim$\rightarrow$Exp see the same number of source-specific mini-batches ($U$ from each source), the same number of source forward/backward evaluations, and the same number $UB_d$ of sample presentations from each source: 60,000 for Cylinder, Controlled Cylinder, and FSI and 30,000 for Foil. They are \emph{source-exposure matched within each dataset}, not optimizer-step matched: Sim$\rightarrow$Exp uses $2U$ optimizer steps, whereas Joint uses $U$. The dataset-specific batch size changes absolute exposure across datasets but not the protocol comparison within a dataset.

\subsection{Optimization and protocol budgets}

Table~\ref{tab:optimization_hparams} lists the optimization settings used in the final runs, including the dataset-specific physical training batch size. Adam uses its default zero weight decay in this code path. Cosine annealing is stepped once after every optimizer update with $T_{\mathrm{max}}=10{,}000$ and $\eta_{\min}=0$ for each stage. Gradient clipping is disabled because the configured clipping threshold is zero.

\begin{table}[H]
\centering
\caption{Optimization settings used in the final runs. The physical training batch size is dataset-specific but fixed across protocols within each dataset.}
\label{tab:optimization_hparams}
\small
\setlength{\tabcolsep}{6pt}
\begin{tabular}{ll}
\toprule
Setting & Value \\
\midrule
Optimizer & Adam \\
Learning rate & $10^{-4}$ \\
Weight decay & 0 \\
Scheduler & cosine annealing to zero, $T_{\mathrm{max}}=10{,}000$ per stage \\
Stage budget & 10,000 optimizer updates \\
Physical train batch size & 6 (Cylinder, Controlled Cylinder, FSI); 3 (Foil) \\
Validation/test batch size & 1 \\
Gradient accumulation & 1 \\
Gradient clipping & disabled \\
Joint loss weights & $0.5$ simulation $+$ $0.5$ experiment \\
Automatic mixed precision & enabled (FP16 autocast with gradient scaling) \\
Data-loader workers & 4 \\
Validation checks & 4 per stage \\
Validation subset & at most 16 experimental batches per check \\
Seed reported & 0 \\
\bottomrule
\end{tabular}
\end{table}

Within a fixed dataset and capacity, the four protocols differ only in how source losses are presented to the optimizer. Their exact budgets are summarized in Table~\ref{tab:protocol_budgets}. A Joint update first backpropagates $0.5\mathcal{L}_{\mathrm{sim}}$ from one simulation mini-batch and then $0.5\mathcal{L}_{\mathrm{exp}}$ from one experimental mini-batch; both gradients are applied by a single optimizer step. The two source computation graphs are not held simultaneously, which permits their native resolutions and field counts to differ.

\begin{table}[H]
\centering
\caption{Optimizer-step and source-exposure budgets. Within each dataset, Joint and Sim$\rightarrow$Exp are matched in source-specific mini-batch and sample exposure and in forward/backward evaluations, but not in optimizer-step count.}
\label{tab:protocol_budgets}
\small
\setlength{\tabcolsep}{5pt}
\resizebox{\linewidth}{!}{%
\begin{tabular}{lrrrl}
\toprule
Protocol & Optimizer steps & Sim. batches & Exp. batches & Stage structure \\
\midrule
Simulation only & 10,000 & 10,000 & 0 & one simulation stage \\
Experiment only & 10,000 & 0 & 10,000 & one experimental stage \\
Sim$\rightarrow$Exp & 20,000 & 10,000 & 10,000 & 10k sim $+$ 10k exp \\
Joint & 10,000 & 10,000 & 10,000 & paired losses before each step \\
\bottomrule
\end{tabular}}
\end{table}

For Sim$\rightarrow$Exp, the experimental stage starts from the best checkpoint produced by that run's simulation stage. Before the experimental stage, Adam, the cosine scheduler, and the mixed-precision gradient scaler are reinitialized; only the learned network weights are transferred. This is the reason the sequential protocol contains 20,000 optimizer steps while Joint contains 10,000.

\subsection{Validation, checkpoint selection, and randomness}

Validation is performed at stage updates 2,500, 5,000, 7,500, and 10,000. Each check evaluates at most 16 batches from the \emph{experimental validation split}; the checkpoint with the lowest experimental-validation RMSE is retained. This same rule is applied to Simulation only, Experiment only, Joint, and both stages of Sim$\rightarrow$Exp. In particular, the simulation-stage checkpoint used to initialize Sim$\rightarrow$Exp is selected by experimental-validation RMSE. The numerical test split is never used for model selection. As noted in the main paper, this uniform rule favors the experimental side and is a limitation of the present study.

For seed 0, Python, NumPy, PyTorch, and all visible CUDA generators are seeded. Loader construction and normalizer setup can consume random numbers, so the global seed is reset immediately before model construction. This ensures identical initial network parameters across the four training modes for a fixed dataset and capacity. Separate deterministic loader generators use offsets $+11$, $+29$, $+37$, $+41$, and $+53$ from the base seed for simulation training, experimental training, experimental validation, experimental test, and numerical test loaders, respectively; worker processes derive their NumPy and Python seeds from the PyTorch worker seed.

\subsection{Test metrics and common-field evaluation}

Every final fluid checkpoint is evaluated independently on the experimental test loader and the numerical test loader. The evaluator generates the entire target block autoregressively, converts predictions back to raw units, concatenates all test predictions, and then calls the public RealPDEBench \texttt{eval\_metrics} implementation. The implementation computes RMSE, MAE, relative $L_2$, $R^2$, kinetic-energy error, fRMSE and its frequency-band variants, and temporal frequency error. The paper reports RMSE, relative $L_2$, and fRMSE in the detailed table and derives bRMSE, bRelL2, and bfRMSE only after the two domain evaluations have been computed separately.

For the four fluid datasets, the evaluator explicitly restricts \emph{both} numerical and experimental predictions to indices $(0,1)$, corresponding to $(u,v)$, before metric computation. Numerical pressure is therefore trained as an additional field but is not included in the primary dual-domain score. Appendix~\ref{app:pressure_retention} separately evaluates pressure on the numerical test split as a missing-field retention test. Because pressure is never observed or supervised in the experimental data, this auxiliary analysis compares the two adaptation strategies, Sim$\rightarrow$Exp and Joint, and uses Simulation only as a reference for pressure performance without experimental adaptation.

\newpage
\section{Extended Results}
\label{app:detailed_results}
\begin{table}[H]
\centering
\caption{Full dual-domain test performance on the four fluid datasets. Both simulation and experimental metrics are computed on the shared velocity fields $(u,v)$. Lower is better. Bold marks the best protocol for each model capacity, test domain, and metric. Results are seed 0.}
\label{tab:dual_domain_metrics}
\scriptsize
\setlength{\tabcolsep}{3.0pt}
\resizebox{\textwidth}{!}{%
\begin{tabular}{llll ccc ccc}
\toprule
& & & & \multicolumn{3}{c}{Simulation test} & \multicolumn{3}{c}{Experimental test} \\
\cmidrule(lr){5-7}\cmidrule(lr){8-10}
Dataset & Model & Params. & Training
& RMSE $\downarrow$ & Rel. $L_2$ $\downarrow$ & fRMSE $\downarrow$
& RMSE $\downarrow$ & Rel. $L_2$ $\downarrow$ & fRMSE $\downarrow$ \\
\midrule
\multirow{8}{*}{Cylinder}
& \multirow{4}{*}{MORPH-Ti} & \multirow{4}{*}{17.5M} & Simulation only
& \textbf{0.014863} & \textbf{0.089989} & \textbf{0.002119}
& 0.166353 & 0.989927 & 0.028099 \\
& & & Experiment only
& 0.108947 & 0.551503 & 0.017663
& 0.093546 & 0.113501 & 0.014074 \\
& & & Sim$\rightarrow$Exp
& 0.105315 & 0.730744 & 0.017025
& 0.095457 & \textbf{0.107778} & 0.014283 \\
& & & Joint
& 0.017445 & 0.107683 & 0.002650
& \textbf{0.080744} & 0.144381 & \textbf{0.013156} \\
\cmidrule(lr){2-10}
& \multirow{4}{*}{MORPH-S} & \multirow{4}{*}{48.2M} & Simulation only
& 0.015774 & 0.085703 & 0.002150
& 0.135024 & 0.770230 & 0.022416 \\
& & & Experiment only
& 0.097461 & 0.520825 & 0.015237
& 0.091060 & 0.103071 & 0.014152 \\
& & & Sim$\rightarrow$Exp
& 0.095836 & 0.607444 & 0.015166
& 0.092475 & \textbf{0.100925} & 0.014120 \\
& & & Joint
& \textbf{0.012276} & \textbf{0.072127} & \textbf{0.001843}
& \textbf{0.078568} & 0.124704 & \textbf{0.012704} \\
\midrule
\multirow{8}{*}{\makecell[l]{Controlled\\Cylinder}}
& \multirow{4}{*}{MORPH-Ti} & \multirow{4}{*}{17.5M} & Simulation only
& \textbf{0.031432} & 0.222405 & 0.007029
& 0.063182 & 0.445163 & 0.016392 \\
& & & Experiment only
& 0.083038 & 0.625413 & 0.020391
& 0.023569 & 0.185535 & 0.004487 \\
& & & Sim$\rightarrow$Exp
& 0.099397 & 0.835142 & 0.025213
& \textbf{0.018297} & \textbf{0.145192} & \textbf{0.003348} \\
& & & Joint
& 0.033027 & \textbf{0.220636} & \textbf{0.006687}
& 0.027616 & 0.212237 & 0.005556 \\
\cmidrule(lr){2-10}
& \multirow{4}{*}{MORPH-S} & \multirow{4}{*}{48.2M} & Simulation only
& \textbf{0.019874} & 0.138686 & 0.004597
& 0.065668 & 0.458582 & 0.016983 \\
& & & Experiment only
& 0.080958 & 0.633147 & 0.019961
& 0.019310 & 0.154054 & 0.003761 \\
& & & Sim$\rightarrow$Exp
& 0.109063 & 0.845359 & 0.027678
& \textbf{0.014985} & \textbf{0.118572} & \textbf{0.002580} \\
& & & Joint
& 0.020592 & \textbf{0.137896} & \textbf{0.004436}
& 0.020318 & 0.160144 & 0.003906 \\
\midrule
\multirow{8}{*}{FSI}
& \multirow{4}{*}{MORPH-Ti} & \multirow{4}{*}{17.5M} & Simulation only
& \textbf{0.036741} & \textbf{0.245198} & \textbf{0.005167}
& 0.077015 & 0.564958 & 0.012695 \\
& & & Experiment only
& 0.114665 & 0.769586 & 0.018818
& 0.021737 & 0.156446 & 0.002545 \\
& & & Sim$\rightarrow$Exp
& 0.096897 & 0.643576 & 0.015303
& \textbf{0.019712} & \textbf{0.142022} & \textbf{0.002116} \\
& & & Joint
& 0.038757 & 0.260137 & 0.005489
& 0.027288 & 0.198263 & 0.003571 \\
\cmidrule(lr){2-10}
& \multirow{4}{*}{MORPH-S} & \multirow{4}{*}{48.2M} & Simulation only
& 0.035331 & 0.236209 & 0.005139
& 0.093100 & 0.676224 & 0.015544 \\
& & & Experiment only
& 0.101363 & 0.651619 & 0.016057
& 0.021288 & 0.152383 & 0.002607 \\
& & & Sim$\rightarrow$Exp
& 0.088639 & 0.560375 & 0.013276
& \textbf{0.018042} & \textbf{0.130585} & \textbf{0.001951} \\
& & & Joint
& \textbf{0.030764} & \textbf{0.209767} & \textbf{0.004252}
& 0.025529 & 0.187325 & 0.003195 \\
\midrule
\multirow{8}{*}{Foil}
& \multirow{4}{*}{MORPH-Ti} & \multirow{4}{*}{17.5M} & Simulation only
& \textbf{0.023643} & \textbf{0.099677} & \textbf{0.003169}
& 0.131021 & 0.584074 & 0.020018 \\
& & & Experiment only
& 0.106668 & 0.499206 & 0.016051
& 0.022516 & 0.060904 & 0.003317 \\
& & & Sim$\rightarrow$Exp
& 0.087303 & 0.457457 & 0.013954
& \textbf{0.018673} & \textbf{0.035981} & \textbf{0.002705} \\
& & & Joint
& 0.028703 & 0.133475 & 0.004086
& 0.023535 & 0.055015 & 0.003810 \\
\cmidrule(lr){2-10}
& \multirow{4}{*}{MORPH-S} & \multirow{4}{*}{48.2M} & Simulation only
& \textbf{0.021560} & \textbf{0.090109} & \textbf{0.003010}
& 0.107166 & 0.462861 & 0.016545 \\
& & & Experiment only
& 0.111847 & 0.495553 & 0.016638
& 0.048071 & 0.140799 & 0.006847 \\
& & & Sim$\rightarrow$Exp
& 0.108193 & 0.563178 & 0.016383
& \textbf{0.024238} & \textbf{0.046342} & \textbf{0.003399} \\
& & & Joint
& 0.042614 & 0.171252 & 0.006316
& 0.024977 & 0.062909 & 0.004017 \\
\bottomrule
\end{tabular}}
\end{table}

\newpage
\section{Sensitivity to Simulation--Experiment Evaluation Weighting.} \label{app:lambda}
To test whether the balanced-performance conclusion depends on the equal-domain choice $\lambda=0.5$, we recompute the dual-domain metrics for evaluation weights $\lambda=0,0.1,\ldots,1$, while keeping the trained checkpoints fixed. Here, $\lambda=0$ corresponds to purely experimental evaluation, $\lambda=1$ to purely simulation evaluation, and $\lambda=0.5$ recovers the balanced metrics in Table~\ref{tab:balanced_dual_domain}. Joint achieves the lowest error in all 24 dataset--capacity--metric comparisons for every $\lambda\in[0.2,0.8]$ sampled here, and remains best in 22/24 comparisons at $\lambda=0.1$ and 21/24 at $\lambda=0.9$. Thus, its advantage is robust over a broad range of domain
preferences and is lost mainly near the single-domain extremes.

\begin{figure*}[h]
\centering
\includegraphics[width=1.0\textwidth,trim=0cm 0cm 0cm 0cm, clip]{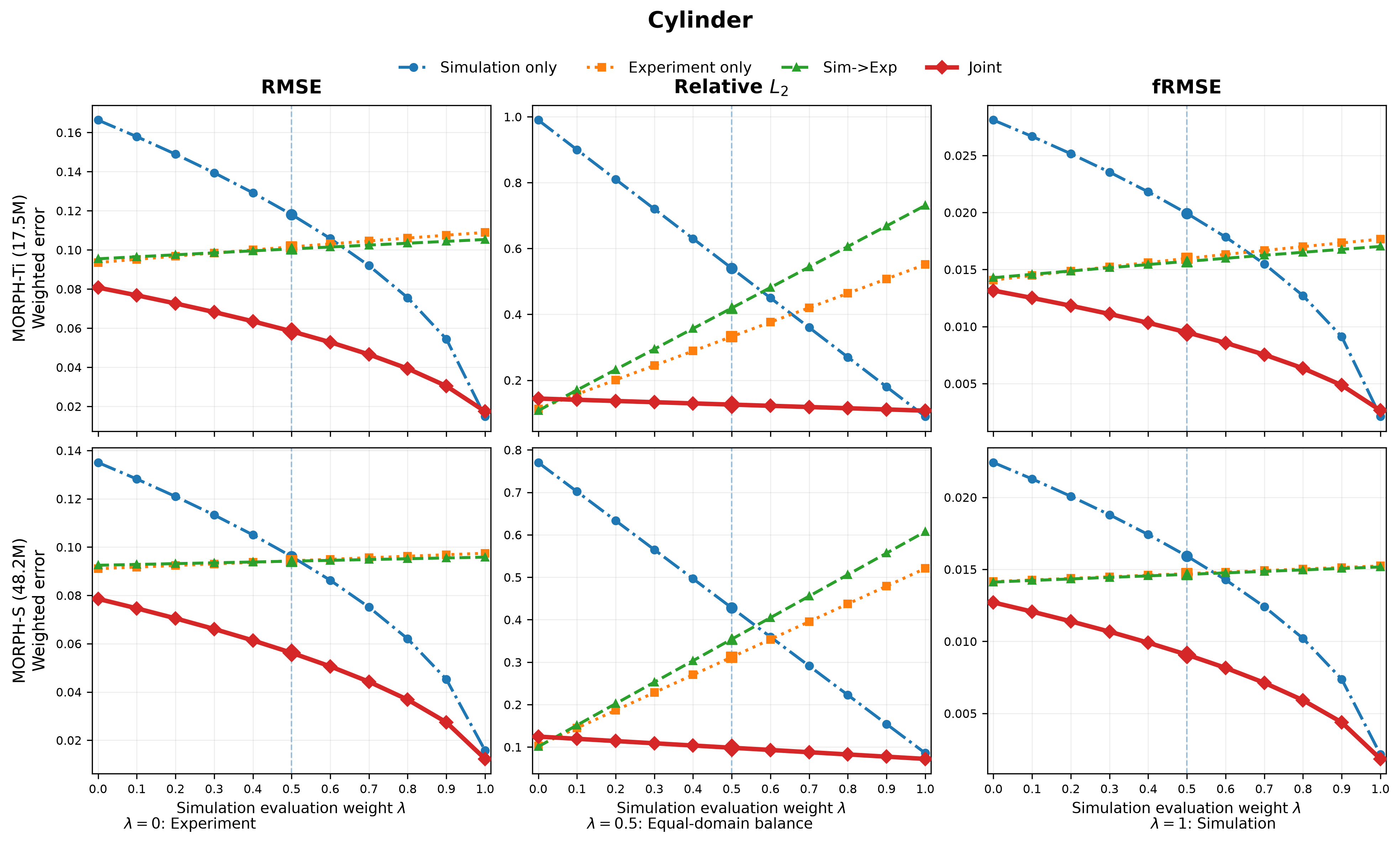}
\caption{\textbf{Cylinder: sensitivity to simulation--experiment evaluation weighting.} Rows show MORPH-Ti and MORPH-S and columns show RMSE, relative $L_2$, and fRMSE (lower is better). Here, $\lambda=0$ weights only experimental performance and $\lambda=1$ only simulation performance; $\lambda=0.5$ recovers the balanced metrics in Table~\ref{tab:balanced_dual_domain}. Joint gives the lowest error for all three metrics and both capacities throughout $\lambda=0.1$--$0.9$. At the single-domain endpoints, Sim$\rightarrow$Exp is lowest only for experimental relative $L_2$, while Simulation only becomes best for MORPH-Ti at $\lambda=1$; Joint remains best for MORPH-S.}
\label{fig:lambda_cylinder}
\end{figure*}

\begin{figure*}[h]
\centering
\includegraphics[width=1.0\textwidth,trim=0cm 0cm 0cm 0cm, clip]{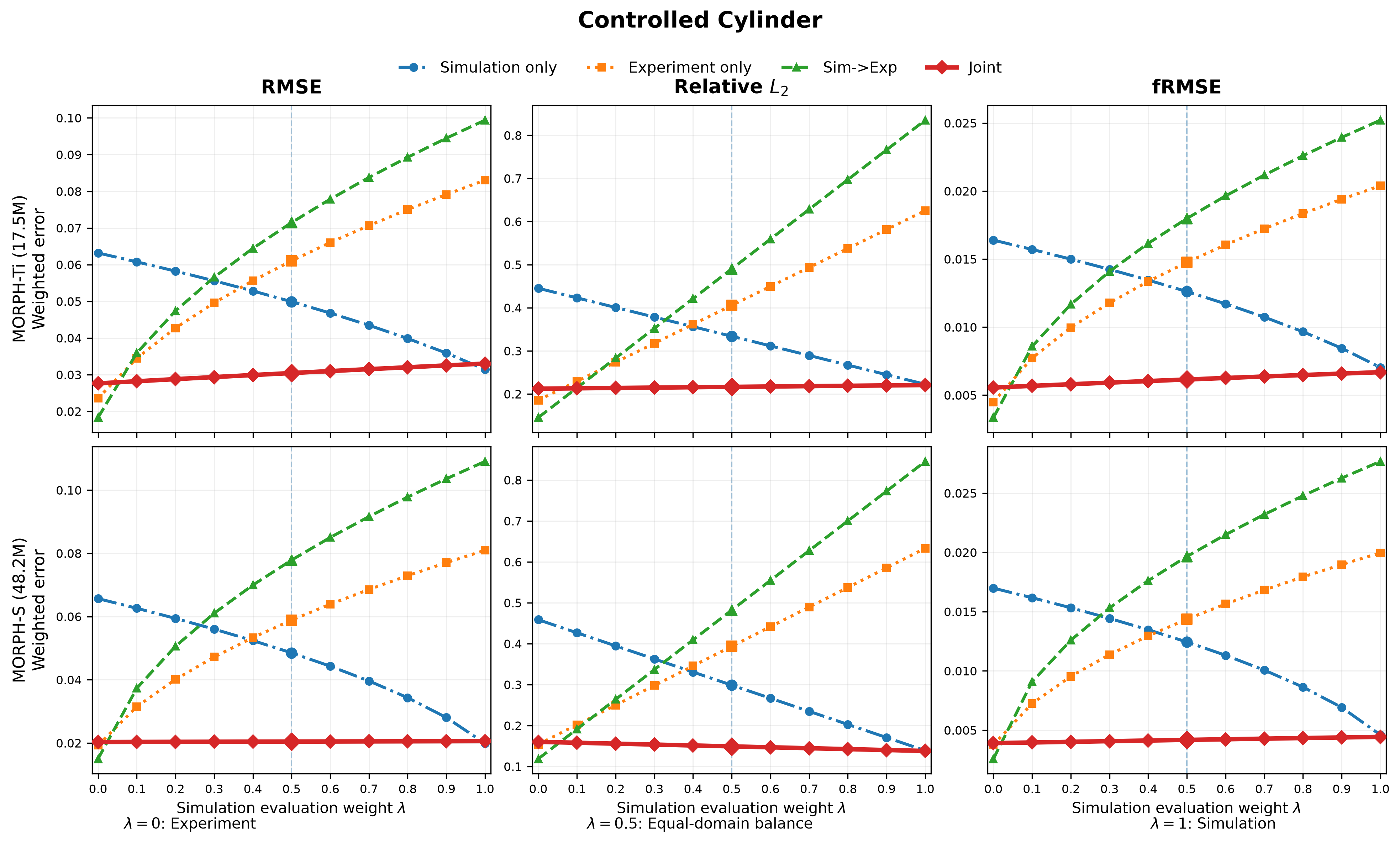}
\caption{\textbf{Controlled Cylinder: sensitivity to simulation--experiment evaluation weighting.} Rows show MORPH-Ti and MORPH-S and columns show RMSE, relative $L_2$, and fRMSE (lower is better). Joint gives the lowest error for all three metrics and both capacities throughout $\lambda=0.1$--$0.9$. Sim$\rightarrow$Exp is best at the purely experimental endpoint ($\lambda=0$), whereas at $\lambda=1$ Simulation only gives the lowest RMSE and Joint retains the lowest relative $L_2$ and fRMSE.}
\label{fig:lambda_ctrlcylinder}
\end{figure*}

\begin{figure*}[h]
\centering
\includegraphics[width=1.0\textwidth,trim=0cm 0cm 0cm 0cm, clip]{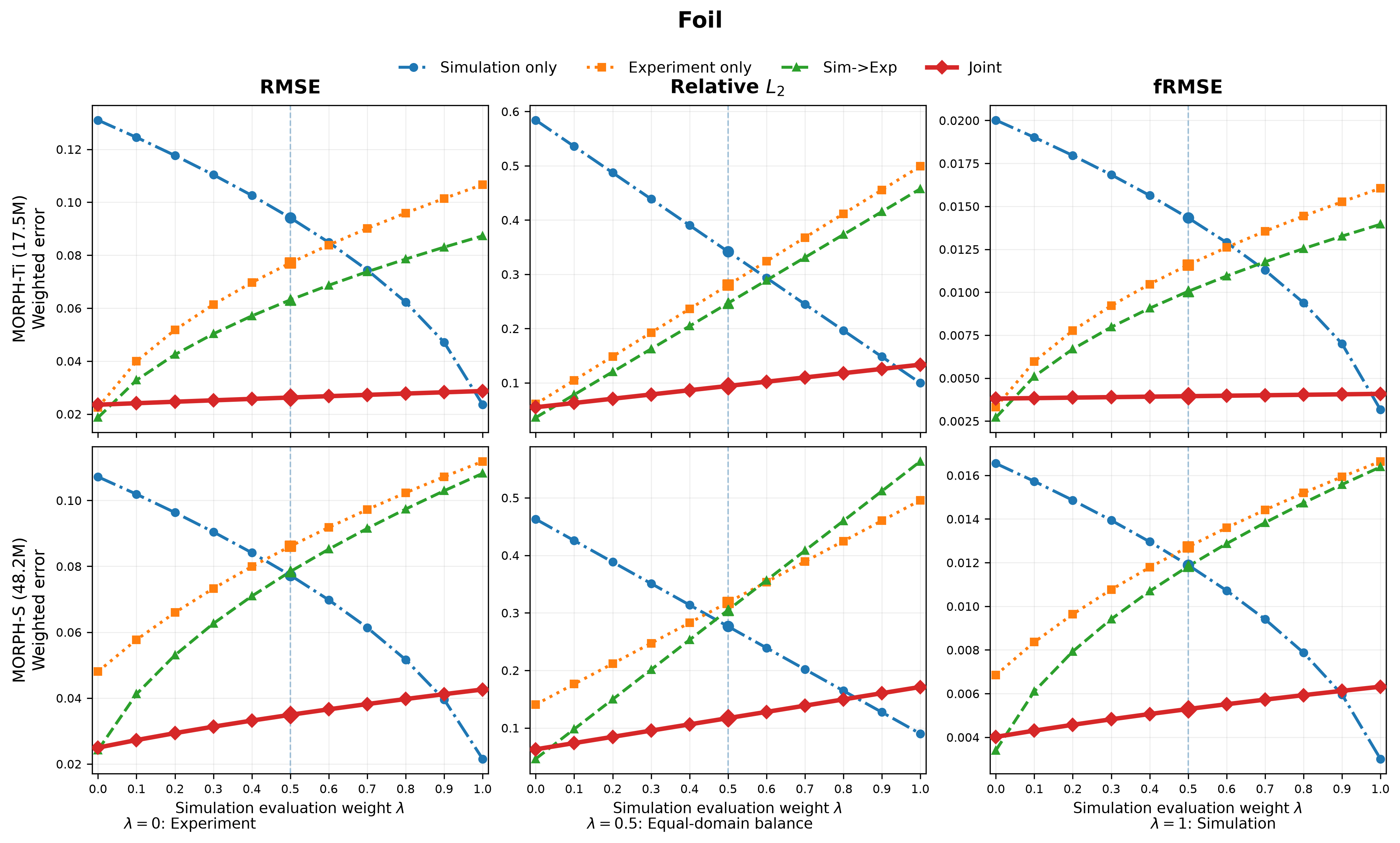}
\caption{\textbf{Foil: sensitivity to simulation--experiment evaluation weighting.} Rows show MORPH-Ti and MORPH-S and columns show RMSE, relative $L_2$, and fRMSE (lower is better). Joint gives the lowest error for all three metrics over $\lambda=0.1$--$0.9$ for MORPH-Ti and $\lambda=0.1$--$0.8$ for MORPH-S. Sim$\rightarrow$Exp is best at the purely experimental endpoint ($\lambda=0$), while Simulation only becomes best near the purely simulation endpoint, at $\lambda=1$ for MORPH-Ti and $\lambda\geq0.9$ for MORPH-S.}
\label{fig:lambda_foil}
\end{figure*}

\begin{figure*}[h]
\centering
\includegraphics[width=1.0\textwidth,trim=0cm 0cm 0cm 0cm, clip]{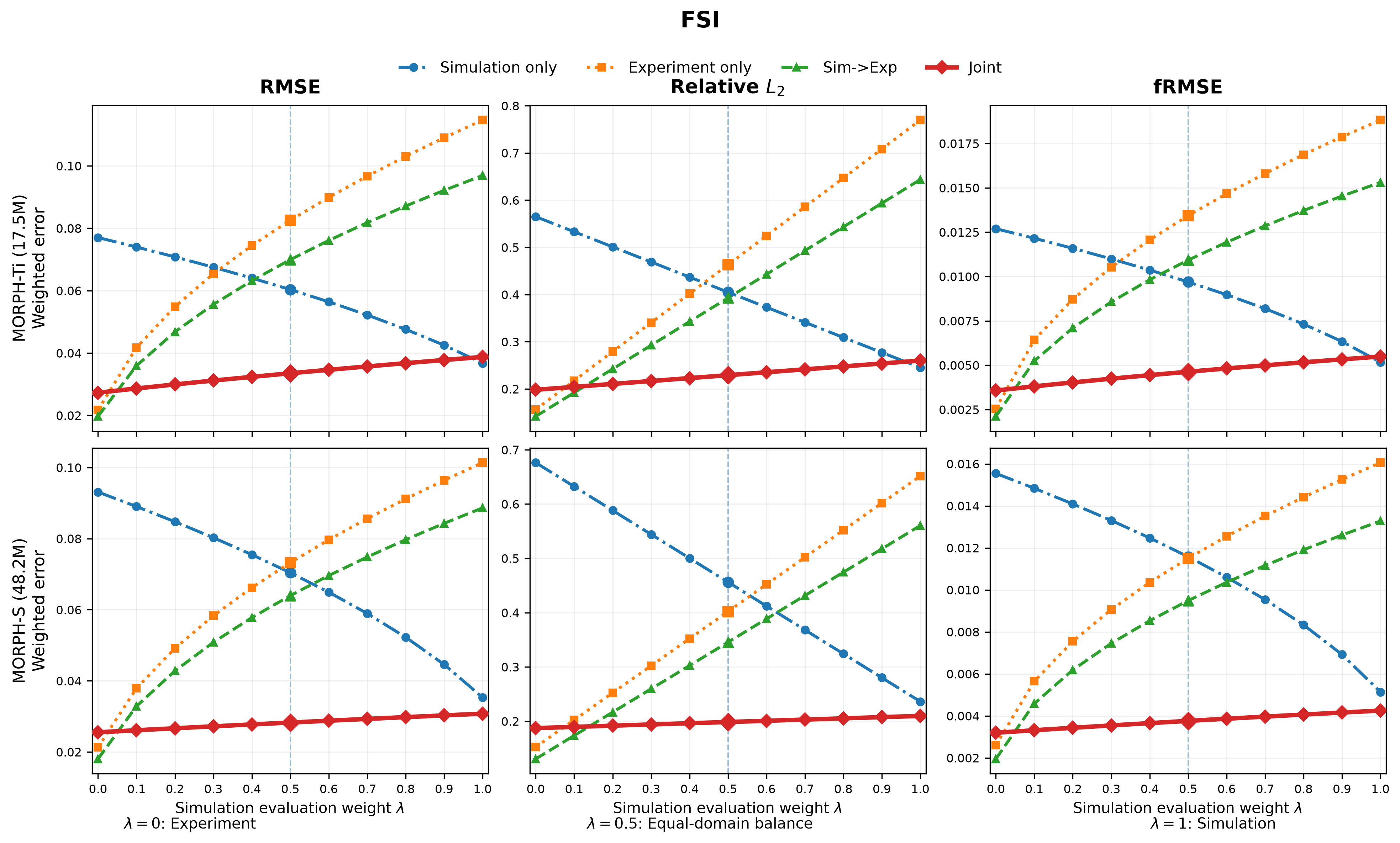}
\caption{\textbf{FSI: sensitivity to simulation--experiment evaluation weighting.} Rows show MORPH-Ti and MORPH-S and columns show RMSE, relative $L_2$, and fRMSE (lower is better). Joint gives the lowest error for all three metrics and both capacities throughout $\lambda=0.2$--$0.9$. At $\lambda=0.1$, Joint remains best for RMSE and fRMSE while Sim$\rightarrow$Exp has the lowest relative $L_2$. Sim$\rightarrow$Exp is best at $\lambda=0$; at $\lambda=1$, Simulation only is best for MORPH-Ti, whereas Joint remains best for MORPH-S.}
\label{fig:lambda_dsi}
\end{figure*}

\clearpage
\section{Missing-Field Retention During Experimental Adaptation}
\label{app:pressure_retention}

Pressure field provides a stricter test of simulation-domain retention than the shared velocity fields because it is present and supervised in the numerical data, but absent from the experimental measurements. During the experimental stage of Sim$\rightarrow$Exp, the model is optimized only on the experimentally observed $(u,v)$ fields, so pressure receives no direct supervision. Joint training, in contrast, continues to receive a numerical mini-batch at every update and therefore retains direct pressure supervision while adapting to the experimental domain.

Table~\ref{tab:pressure_retention} compares the two adaptation strategies on numerical-pressure prediction. Joint substantially outperforms Sim$\rightarrow$Exp in every dataset--capacity setting: it reduces pressure RMSE
by 52.4--86.6\% (median 61.6\%), relative $L_2$ by 54.1--87.0\% (median 60.5\%), and fRMSE by 58.8--86.9\% (median 64.8\%). Thus, Joint is better than Sim$\rightarrow$Exp in all 24 pressure-metric comparisons (8 settings $\times$ 3 metrics).

These results expose a particularly clear form of simulation-domain forgetting. Once Sim$\rightarrow$Exp switches to experimental fine-tuning, pressure is no longer part of the supervised objective, and the learned capability for this
simulation-only field is substantially weakened. Joint avoids this loss by keeping simulation supervision active throughout adaptation. The consistent advantage across physical-space RMSE, relative $L_2$, and Fourier-space RMSE
shows that preserving the missing field is not specific to a single error measure.

\begin{table}[H]
\centering
\caption{Numerical-pressure retention after experimental adaptation. Pressure $p$ is supervised in simulation but is unavailable in the experimental measurements. The table therefore focuses on Sim$\rightarrow$Exp and Joint. Experiment only is not reported because pressure is never supervised in that protocol. Lower is better for the error metrics.
The final column reports the percentage reduction in pressure RMSE achieved by Joint relative to Sim$\rightarrow$Exp,
$100(E_{\mathrm{Sim\rightarrow Exp}}-E_{\mathrm{Joint}})/E_{\mathrm{Sim\rightarrow Exp}}$. Joint is lower than Sim$\rightarrow$Exp in all 24 metric comparisons (8 settings $\times$ 3 metrics).}
\label{tab:pressure_retention}
\scriptsize
\setlength{\tabcolsep}{3.5pt}
\resizebox{\textwidth}{!}{%
\begin{tabular}{ll ccc ccc c}
\toprule
& &
\multicolumn{3}{c}{Sim$\rightarrow$Exp} &
\multicolumn{3}{c}{Joint} &
\multirow{2}{*}{\makecell{RMSE reduction\\(\%) $\uparrow$}} \\
\cmidrule(lr){3-5}\cmidrule(lr){6-8}
Dataset & Model
& RMSE $\downarrow$ & Rel. $L_2$ $\downarrow$ & fRMSE $\downarrow$
& RMSE $\downarrow$ & Rel. $L_2$ $\downarrow$ & fRMSE $\downarrow$
& \\
\midrule

\multirow{2}{*}{Cylinder}
& MORPH-Ti
& 0.292374 & 0.973175 & 0.050192
& \textbf{0.055293} & \textbf{0.179063} & \textbf{0.008347}
& \textbf{81.1} \\

& MORPH-S
& 0.261182 & 0.869020 & 0.040659
& \textbf{0.034885} & \textbf{0.112863} & \textbf{0.005319}
& \textbf{86.6} \\

\midrule

\multirow{2}{*}{\makecell[l]{Controlled\\Cylinder}}
& MORPH-Ti
& 0.262821 & 0.716707 & 0.067997
& \textbf{0.116303} & \textbf{0.315430} & \textbf{0.023091}
& \textbf{55.7} \\

& MORPH-S
& 0.225042 & 0.613608 & 0.050295
& \textbf{0.071026} & \textbf{0.192083} & \textbf{0.013899}
& \textbf{68.4} \\

\midrule

\multirow{2}{*}{FSI}
& MORPH-Ti
& 0.661982 & 0.947801 & 0.102268
& \textbf{0.270900} & \textbf{0.400808} & \textbf{0.042135}
& \textbf{59.1} \\

& MORPH-S
& 0.622418 & 0.907676 & 0.093517
& \textbf{0.223653} & \textbf{0.333589} & \textbf{0.035381}
& \textbf{64.1} \\

\midrule

\multirow{2}{*}{Foil}
& MORPH-Ti
& 0.127712 & 0.981016 & 0.018797
& \textbf{0.060559} & \textbf{0.448919} & \textbf{0.006837}
& \textbf{52.6} \\

& MORPH-S
& 0.122945 & 0.942570 & 0.018282
& \textbf{0.058501} & \textbf{0.432566} & \textbf{0.006784}
& \textbf{52.4} \\

\bottomrule
\end{tabular}}
\end{table}

\section{Limitations}

Joint is trained only with $\lambda=0.5$. Appendix~\ref{app:lambda} varies the evaluation weighting with fixed checkpoints, but training-time sensitivity to $\lambda$ remains unexplored. The gradient analysis is local, formulated before Adam preconditioning, and gradient alignment is not measured directly. The models are also trained with one-step teacher forcing and evaluated autoregressively, so absolute errors should not be interpreted as an architecture-matched reproduction of RealPDEBench baselines that predict complete target blocks.

\end{document}